\documentclass[journal]{IEEEtran}
\ifCLASSINFOpdf
\else
\fi
\usepackage{times}
\usepackage{soul}
\usepackage{url}
\usepackage[hidelinks]{hyperref}
\usepackage[utf8]{inputenc}
\usepackage{graphicx}
\usepackage{amsmath}
\usepackage{amssymb}
\usepackage{booktabs}
\usepackage{algorithm}
\usepackage{algorithmic}
\usepackage{color}
\usepackage{enumerate}
\usepackage{xcolor}
\usepackage[caption=false]{subfig}
\usepackage{pifont}

\def\etal{\emph{et al.,}}

\begin{document}
%
\title{Latent Adversarial Defence with Boundary-guided Generation}
%
%
%

\author{Xiaowei~Zhou,
        Ivor~W.~Tsang,
        and~Jie~Yin,~\IEEEmembership{Member,~IEEE}
\thanks{X. Zhou is with the Centre for Artificial Intelligence, FEIT, University of Technology Sydney, Ultimo, NSW 2007, Australia and Data61, CSIRO, NSW 2122, Australia (e-mail:Xiaowei.Zhou@student.uts.edu.au).}
\thanks{I. W. Tsang is with the Centre for Artificial Intelligence, FEIT, University of Technology Sydney, Ultimo, NSW 2007, Australia
(e-mail: ivor.tsang@uts.edu.au).}
\thanks{J. Yin is with the Discipline of Business Analytics, The University of Sydney, NSW 2006, Australia (e-mail: jie.yin@sydney.edu.au).}
}

%
%

\markboth{Journal of \LaTeX\ Class Files,~Vol.~14, No.~8, August~2015}%
{Zhou \MakeLowercase{\textit{et al.}}: Latent Adversarial Defence with Boundary-guided Generation}
%



\maketitle

\begin{abstract}
Deep Neural Networks (DNNs) have recently achieved great success in many tasks, which encourages DNNs to be widely used as a machine learning service in model sharing scenarios. However, attackers can easily generate adversarial examples with a small perturbation to fool the DNN models to predict wrong labels. To improve the robustness of shared DNN models against adversarial attacks, we propose a novel method called \textit{Latent Adversarial Defence} (LAD). 
The proposed LAD method improves the robustness of a DNN model through adversarial training on generated adversarial examples. Different from popular attack methods which are carried in the input space and only generate adversarial examples of repeating patterns, LAD generates myriad of adversarial examples through adding perturbations to latent features along the normal of the decision boundary which is constructed by an SVM with an attention mechanism. Once adversarial examples are generated, we adversarially train the model through augmenting the training data with generated adversarial examples. Extensive experiments on the MNIST, SVHN, and CelebA dataset demonstrate the effectiveness of our model in defending against different types of adversarial attacks.
\end{abstract}


%
\IEEEpeerreviewmaketitle

\section{Introduction}
\label{introduction}
%
%
%
%

\begin{figure}[htbp]
    \centering
    \includegraphics[scale=0.88]{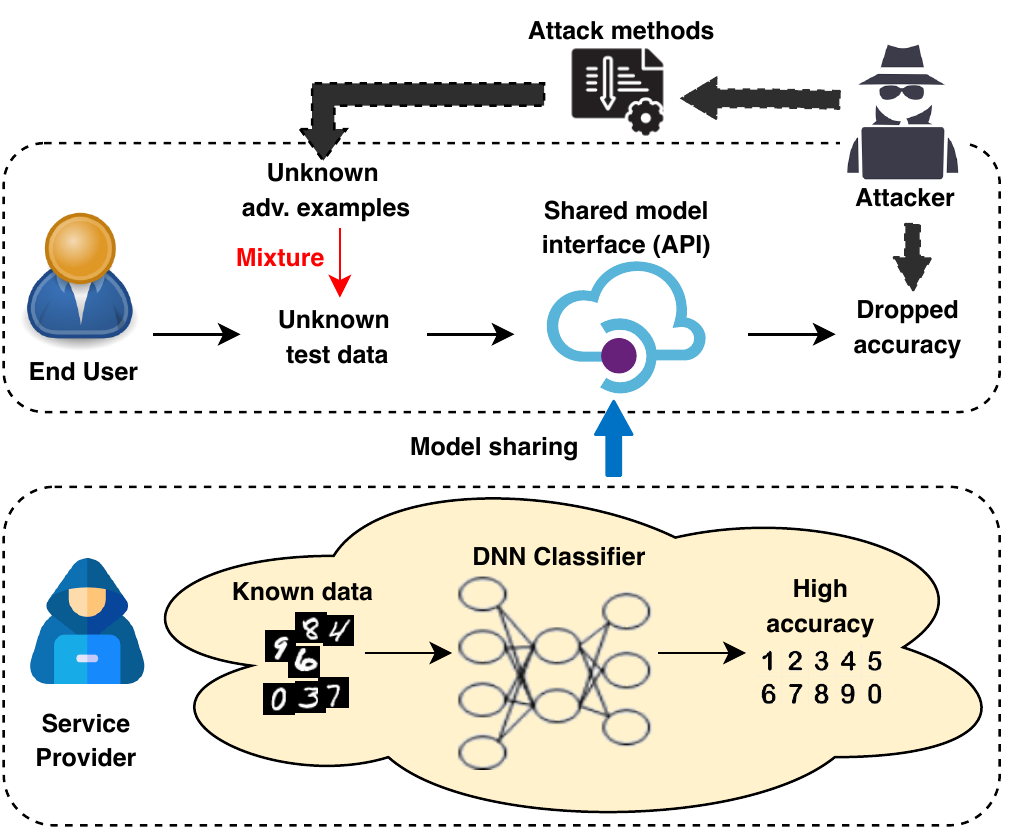}
    \vspace{-0.5cm}
    \caption{Attacks in model sharing scenario. Service provider trains a high-performanced DNN classifier and shares it to public. The end user performs classification task through the shared model without knowledge about the generated adversarial examples which are generated by  attackers to attack the shared model interface.}
    \label{fig:model_share}
\end{figure}

\IEEEPARstart{I}{n} recent years, deep neural network models have been successfully applied in many areas, such as image processing~\cite{huang2017densely,shi2019label}, speech~\cite{seide2011conversational,amodei2016deep} and natural language processing~\cite{fedus2018maskgan,alikaniotis2016automatic}, etc. Training a DNN model often requires large amounts of labeled data and significant efforts of parameter tuning. As such, it accelerates the development of DNN models to be hosted in a cloud and run as a service that can be shared by a third party. This leads to many online machine learning as service platforms (MLaaS) that provide Web-based API services for various tasks based on DNN models. For example, image and video analysis from the AWS pre-trained AI Services~\cite{Amazon}, powerful image analysis from Google Cloud Vision~\cite{Google}. 

In such model sharing scenarios, the increasing use of DNN models, however, has raised serious security and reliability concerns. This can be illustrated in Fig.~\ref{fig:model_share}. The service providers train DNN models using the dataset they collect, which are expected to achieve high classification accuracy on test examples having similar distributions with the training dataset. The trained DNN models can be hosted in the cloud and run as a service. End users may then provide their own test examples, mostly unknown to the service provider in advance, into the shared model to obtain prediction results. However, very often, test examples are very likely to be mixed with unknown adversarial examples~\cite{szegedy2013intriguing,yuan2019adversarial}, which could be unnoticeable to end users. Adversarial examples can be generated by attackers through adding small crafted perturbations to legitimate examples. These examples are often indistinguishable to human eyes, so they can easily fool DNN models to predict wrong labels. What makes it even worse is that these adversarial examples can be generated by various types of unknown attack methods, leading a dramatic drop in classification accuracy. This is attributed to the fact that the shared DNN models are not robustly trained to defend various unknown adversarial attacks before they are released as a service. Thus, in this paper, we are mainly concerned about how to enhance the adversarial robustness of shared DNN models from service provider perspectives.

\begin{figure*}[htbp]
    \centering
    \includegraphics[scale=0.88]{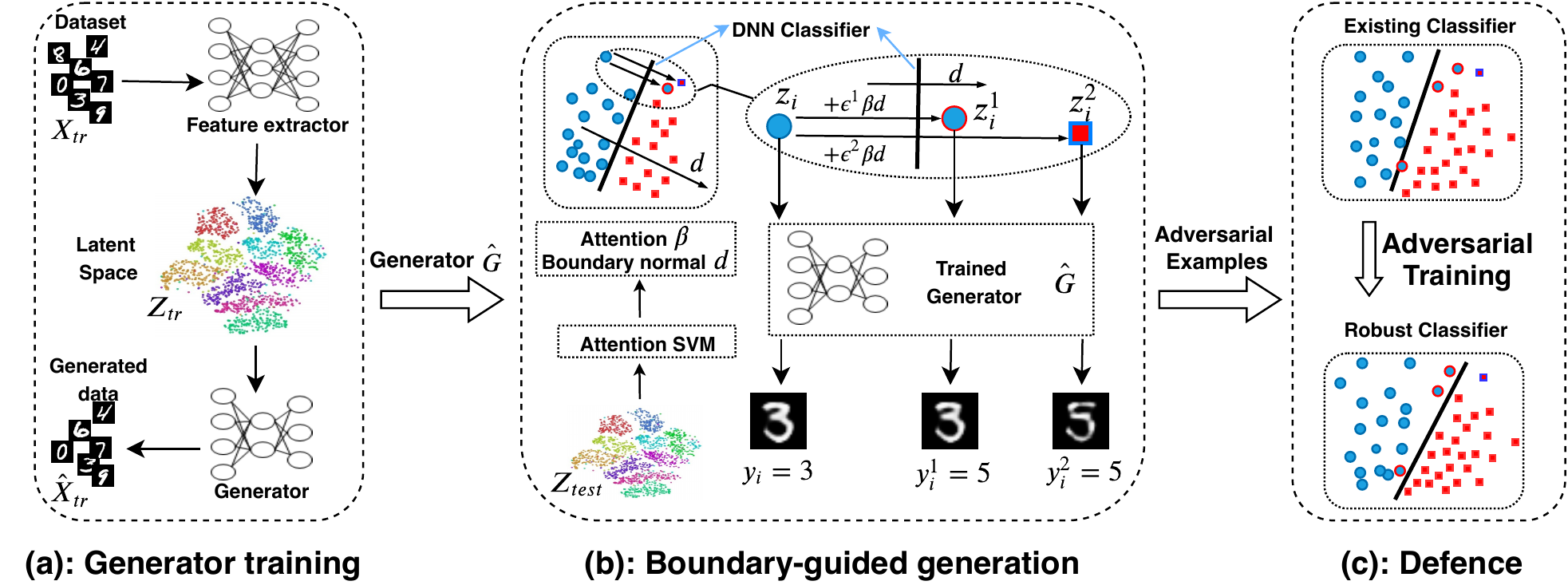}
    \vspace{-0.15cm}
    \caption{Overview of \textit{Latent Adversarial Defence}. (a) Train a generator through feature extractor, i.e., DNN classifier, to decode latent features to images. (b) Generate adversarial examples by perturbing latent features alongside the decision boundary norm of attention SVM. $z_i$ is the latent feature; $\beta$ is attention weights; $d$ is the boundary norm; $y_i$ is the label of the example. (c) Adversarially train the DNN classifier to improve adversarial robustness. The blue dots with red circle are adversarial examples.}
    \label{fig:flowchat}
\end{figure*}

To enhance the robustness of DNN models against various adversarial attacks, adversarial training methods have been shown to be most effective~\cite{Goodfellow2014Exp,shaham2018understanding,shrivastava2017learning}. The key idea is to augment the training set with generated adversarial examples to train the DNN model before it is released as service. Different methods have been proposed to generate adversarial examples through adding a small perturbation to a legitimate input sample~\cite{Goodfellow2014Exp,madry2017towards,papernot2016limitations,carlini2017towards}. Formally, for a given classifier $f$, the predicted label of an input sample $x$ is defined as $f(x)$, and $f(x')$ is the label of adversarial example $x'$: 
\begin{equation}
\begin{aligned}
    x' = x + \delta, \mbox{ s.t. }
    f(x') \neq f(x), 
    \label{eqa:adv}
\end{aligned}
\end{equation}
where $\delta$ is the small perturbation added to an input sample $x$. $x'$ is the generated adversarial example in the neighborhood of the original legitimate sample in the input space. The original input examples can be augmented with adversarial examples generated this way to adversarially train the DNN model. 

However, most of the existing methods suffer from two major limitations. First, because the perturbation is added with respect to individual input examples, these methods often only craft adversarial examples of repeating patterns. The DNN models adversarially trained with these examples would be only effective to defend very specific types of adversarial attacks, but still vulnerable to other unknown adversarial attacks. Therefore, this presses the need to increase the diversity of generated adversarial examples so that the DNN model can fully explore the unknown adversarial example space to improve the robustness.

Second, most of the methods are carried out in the input space, where legitimate examples are often corrupted with noise and have complex distributions. As such, operating in the input space may mislead the generation of adversarial examples, which would severely hurt the DNN model. Only recently, Song \emph{et al.}~\cite{song2018constructing} proposed a deep learning based attack method, GA, to generate adversarial images in the latent space. They explored the AC-GAN~\cite{odena2017conditional} latent space to generate adversarial images that would most likely mislead the targeted classifier. However, the GA method treats all examples equally, but neglects the decision boundary information while generating adversarial examples. This inevitably reduces the effectiveness of adversarial training. In essence, the decision boundary plays a critical role in guiding the generation of adversarial examples. Intuitively, adversarial examples close to the decision boundary would best contribute to enhancing the DNN model, while those far away from the decision boundary would be much less useful. However, how to use the decision boundary to guide the generation of adversarial examples is less explored.

To tackle aforementioned limitations, in this paper, we propose a novel method called \textit{Latent Adversarial Defence} (LAD), which is designed to generate myriad of adversarial examples based on the decision boundary in a latent space and to robustly defend against different unknown adversarial attacks. Fig.~\ref{fig:flowchat} gives an overview of the proposed LAD model. Unlike the existing methods that operate in the input space, LAD generates adversarial examples by adding perturbations to latent features, which can be used to construct the decision boundary more accurately. Our adversarial examples generation is guided by the normal of decision boundary in latent space, which is learned through a support vector machine (SVM)~\cite{boser1992training} with an attention mechanism. This provides an unrestricted way to generate a variety of adversarial examples. After adversarial training on generated adversarial examples, the fine-tuned DNN model is effective in defending against different types of unknown adversarial attacks. Comprehensive experiments are conducted on the MNIST, SVHN, and CelebA dataset to verify the effectiveness of the proposed model. 

The contribution of this paper is three-fold:
\begin{itemize}
    \item We propose a new method for generating adversarial examples by adding boundary guided perturbations in the latent feature space, which are inherently different from the forms of the original examples.
    \item Our generator can exhibit transitions of adversarial examples in both feature and label space across the decision boundary, providing valuable insights regarding when and how potential attacks happen.
    \item After adversarial training on our generated adversarial examples,  the fine-tuned model improves its adversarial robustness and better defends different types of attacks.
\end{itemize}

\section{Related Work}
This section reviews two main branches of related literature: adversarial attack methods and adversarial defence approaches.

\subsection{Adversarial Attack}


From methodology point of view, existing adversarial attack methods can be grouped into two categories: \textit{gradient-based methods} and \textit{label-based methods}. Gradient-based attack methods modify an input sample in the direction of the gradient of loss function with respect to the input sample. Goodfellow \etal~\cite{Goodfellow2014Exp} proposed a gradient based attack approach, called fast gradient sign method (FGSM), which uses the sign of gradient ($\nabla_{x}J(\theta, x, y)$) of loss function with respect to input examples as perturbation. 
Built upon FGSM, the one-step attack method, Madry \emph{et al.}, \cite{madry2017towards} proposed a multi-step attack method called PGD. PGD iteratively uses the gradient information and generates adversarial examples on the results of the last step. Papernot \emph{et al.,}~\cite{papernot2016limitations} introduced saliency map based on Jacobian matrix into the generation of adversarial examples. The saliency values computed by forward derivative of a target model are used as an indicator to determine the locations in input examples to add perturbation. This method is called Jacobian saliency map attack (JSMA). 

Label-based methods, on the other hand, manipulate the labels of training  data to make the learned DNN model beneficial to their specific purposes. This line of methods use Eq. (\ref{eqa:adv}) as a measure, i.e., changing the original label to target label, to generate adversarial examples. Carlini \emph{et al.} proposed an approach, CW~\cite{carlini2017towards}, to generate adversarial examples by adding small change on the original images in the input space. CW tries to minimize the distance between benign examples and adversarial ones, while enforcing the label of adversarial examples as the targeted ones. A deep learning based attack method, GA, was developed by Song \emph{et al.}~\cite{song2018constructing}. GA exploits an AC-GAN~\cite{odena2017conditional} to generate adversarial images via exploring the latent space of random noise input $z$. 


We summarize and compare different methods for generating adversarial examples, as shown in Table~\ref{tab:attack_methods}. Most of the existing attack methods are carried in the input space and can only craft adversarial examples of repeating patterns. Although GA~\cite{song2018constructing} attempts to generate adversarial examples in the latent space, the crucial information on the decision boundary is not considered. To fill the gap, our LAD method generates adversarial examples through perturbing latent features, which can greatly increase the diversity of adversarial examples through inverting the perturbed latent features to input space. Guided by the decision boundary, the proposed LAD model is able to generate a diversity of adversarial examples near the decision boundary (the most uncertain examples) to improve robustness of DNN models through adversarial training.


\begin{table}[tbp]
\centering
\caption{Comparison of different methods for generating adversarial examples.}
\resizebox{0.51\textwidth}{13mm}{
\setlength{\tabcolsep}{0.7mm}{
\begin{tabular}{cccccc}  
\toprule
Attack methods & Gradient-based & Label-based & Input space & Latent space & Boundary\\ 
\midrule
FGSM~\cite{Goodfellow2014Exp} & \checkmark & \ding{55} & \checkmark & \ding{55}  & \ding{55}\\
PGD~\cite{madry2017towards} & \checkmark & \ding{55} & \checkmark & \ding{55} & \ding{55} \\
JSMA~\cite{papernot2016limitations} & \checkmark & \ding{55} & \checkmark & \ding{55} & \ding{55} \\
CW~\cite{carlini2017towards} & \ding{55} & \checkmark & \checkmark & \ding{55} & \ding{55} \\
GA~\cite{song2018constructing} & \ding{55} & \checkmark & \ding{55} & \checkmark & \ding{55} \\
LAD (ours) & \ding{55} & \checkmark & \ding{55} & \checkmark & \checkmark \\
\bottomrule
\end{tabular}
}
}
\label{tab:attack_methods}
\end{table}

\subsection{Adversarial Defence} 
For various types of adversarial attacks, a key research question is, how can one improve the adversarial robustness of a DNN model before it is deployed as a service? In response, defence strategies have been proposed to mitigate the effect of adversarial attacks. 
Adversarial defence methods can be generally divided into three categories: 1) \textit{Defensive distillation}~\cite{papernot2016distillation,papernot2017extending}, which aims at learning a smooth targeted defence model with $d$ training times. 
The ground truth labels are used as supervised information for the first trained model. Except for the first time, the predicted labels from $(d-1)$-th model are used as ground truth to train the $d$-th model. Finally, the $d$-th model $F^d$ is the distilled model, which is more robust against adversarial attacks. 2) \textit{adversarial example de-noising}, which tries to remove the perturbation added on adversarial examples to obtain a similar sample to the original one. The de-nosing examples can be easily processed by the model, compared with adversarial examples. Defence-GAN~\cite{samangouei2018defense} and MagNet~\cite{meng2017magnet} are two typical methods falling into this category. 3) \textit{Adversarial training}~\cite{Goodfellow2014Exp,shaham2018understanding,shrivastava2017learning} is an effective defence method, which defends attacks by augmenting the training data with adversarial examples when training the targeted model. Adversarial training can be carried either by training the targeted model with the original and adversarial examples ~\cite{kurakin2016adversarial} or with a modified loss function~\cite{Goodfellow2014Exp}. 

Among these methods, defensive distillation and adversarial example de-noising methods may incur high computational overhead, because it is computationally expensive to additionally train the shared models several times or de-noise the adversarial examples by a GAN before classification. This makes them unsuitable to be deployed for model sharing scenarios. Our proposed LAD model falls into adversarial training based defence methods. Unlike the existing methods that are only effective to specific attacks, LAD generates a variety of adversarial examples guided by the decision boundary in the latent space. Therefore, it is effective to robustly defend different types of unknown adversarial attacks.




\section{Latent Adversarial Defence}

To enhance adversarial robustness of DNN models, \textit{Latent adversarial defence} (LAD) model aims to generate diverse adversarial examples based on a decision boundary constructed in a latent space. From the perspective of the latent space, perturbations are added to latent features, which is guided by the normal of decision boundary. To observe the corresponding changes in the input space, we train a generator to map the latent features back to the input space. When the strength of perturbation increases, the corresponding adversarial examples in the input space towards the target label can be generated by the trained generator. Through adversarial training, generated adversarial examples are used to improve the adversarial robustness of DNN models. 

In the following, we present the details of our proposed LAD model.

\subsection{Training of Boundary-guided Generator }
\label{sec:generatrion}
As mentioned above, the generator is trained to invert latent features to the input space, so that humans can understand what changes happen in the input space, caused by changes of latent features. 

For a specific DNN model such as LeNet-5, we learn a generator $\hat{G}$ which is trained on a dataset $X_{tr}=\{x_1,...,x_n\}$ to map latent features to the input space. As shown in Fig.~\ref{fig:flowchat} (a), each sample in the training dataset is fed into the feature extractor, i.e., the DNN model to extract the corresponding latent features. The output of any fully connected layer of a DNN model can be used to construct the latent space $Z_{tr}=\{z_1,...,z_n\}$. Sample $x_i$ and its corresponding latent features $z_i$ are fed in the designed generator. 
The objective function of $G$ over the neural network class $\mathcal{G}$ is as follows:
\begin{equation}
    \hat G =\arg \min\limits_{G\in\mathcal{G}} n^{-1} \sum_{i=1}^{n}\left\|x_{i} - G(z_i)\right\|_p^p,
    \label{eqa:gene}
\end{equation}
where $\|\cdot\|_p$ represents $l_p$ norm, $p=1~or~p = 2$ in this paper; $z_i=\phi(x_{i})$, where $\phi$ is a feature extractor, part of a DNN model.

A mapping between the latent space and the input space is learned by optimizing Eq. (\ref{eqa:gene}). Thus, reconstructed data point $\hat{x_i}$ can be obtained by passing $z_i$ to the trained generator $\hat{G}$. 

\subsection{Latent Boundary-guided Generation} \label{sec:attack}

We generate adversarial examples in the latent space, which is guided by the decision boundary obtained from learning a support vector machine (SVM) on latent features. As perturbation strength $\epsilon$ increases, the predicted label of a generated sample would change from the correct one to the false one, while humans may think the label of the generated sample remains unchanged. For example, in Fig.~\ref{fig:flowchat} (b), a small perturbation $\epsilon^1$ is added to latent features $z_i$, labels of the first two generated examples are both $3$ for human eyes, but the targeted DNN model predicts the two images ($y_i$ and $y_i^1$) as $3$ and $5$, respectively.

\paragraph{Boundary-guided Attention} Firstly, in order to obtain the decision boundary between two classes, a linear SVM with attention mechanism is trained on latent features extracted from the test dataset. 
Because the training dataset is usually inaccessible in model sharing scenarios, we use a small number of test examples, usually 20$\sim$200 examples, to construct a test latent space $Z_{test}$ through the targeted DNN model. After constructing the test latent space with the corresponding labels, a linear SVM with attention is trained. Therefore, a decision boundary and its normal are defined at the same time. 

Note that, attention mechanism~\cite{zhang2018self} is used when we train the linear SVM, which is beneficial to capture a better representation with different weights assigned to different parts of latent features. Specifically, an attention layer is added to process latent features, rather than directly taking them as input to SVM. 
The attention layer is defined as follows:
\begin{eqnarray*}
\alpha_i\!=\! \tanh\left(conv(z_i)\right)\!, \, \beta_i^j \!=\!\! \frac{\exp(\alpha_i^j)}{\sum_{j=1}^N \!\exp(\alpha_i^j)}\!,  
\!\!&\!\! \mbox{and}\!\! &\!\! z_i^{att} \!=\! \beta_i  z_i,
\end{eqnarray*}
where $z_i$ is a latent feature; $conv$ represents convolutional operation; $\tanh$ is the activation function; $\beta_i$ is the attention; $z_i^{att}$ is the output after applying attention operation on the latent feature $z_i$.

\paragraph{Latent Feature Perturbation} Secondly, latent features for each sample are modified according to the normal of the decision boundary obtained in the previous step. The normal $d$ provides a direction guiding our generation and the attention $\beta$ captures the importance of different parts of latent features to moving across the boundary. Different perturbations ($\epsilon$) are added to the same latent features $z_i$ to obtain the modified latent features $z_i^1$ and $z_i^2$ by: 
\begin{eqnarray*}
        z_i^1 = z_i + \epsilon^1 \beta_i d &\mbox{ and } &
        z_i^2 = z_i + \epsilon^2 \beta_i d,
\end{eqnarray*}
where $d$ is the normal of decision boundary from the linear SVM; $\beta_i$ is the attention obtained by SVM for each sample; $\epsilon^1$ and $\epsilon^2$ represent different strength of perturbation. To see more clearly, Fig.~\ref{fig:flowchat} (b) illustrates the change of relationship between modified latent features and the DNN model. 
For $z_i^2$, the predicted label of the generated example has changed, but the real label is still the same as the original example.

If $z_i^1$ and $z_i^2$ are further fed into the remaining part of the DNN model, we could obtain the labels of the modified latent features. Here, the remaining part of the DNN model refers to the remaining layers after the layer from which latent features are extracted. We denote this remaining part as $f_{p}$. For example, if we extract features of the second-to-last fully connected layer of LeNet, the remaining part of the DNN model is the last fully connected layer and softmax classifier.

When the value of perturbation is big enough, the label of the modified latent features would change from positive (negative) to negative (positive), which means that it would cross the decision boundary of the DNN model. As shown in Fig.~\ref{fig:flowchat}~(b), modified latent features $z_i^1$ move from the left side of the classifier to the right side. As the strength of perturbation continues to increase, modified latent features $z_i^2$ would move far away from the decision boundary.

\paragraph{Boundary-guided Generation} Thirdly, modified latent features are inverted to examples in the input space, so that humans can understand the changes caused by different strength of perturbation added. As we can see from Fig.~\ref{fig:flowchat} (b), different latent features including the original one $z_i$ are fed into the trained generator $\hat{G}$ to obtain the corresponding reconstructed image $\hat {x_i}$: $\hat {x_i} = \hat G\left(z_{i}+\epsilon  \beta d\right)$, where $\epsilon \geq 0$.

Ideally, the classification results of latent features being directly fed into the remaining DNN model $f_p$ and the corresponding reconstructed image being fed into the DNN model $f$ should be consistent. That is, the following equation should be satisfied:
\begin{equation}
    f_p\left(z_i^j\right) = f\left(\hat{G}(z_i^j)\right).
    \label{eqa:label_equal}
\end{equation}
However, Eq.~(\ref{eqa:label_equal}) does not always hold true for some modified latent features. For those examples, whose labels of latent features and reconstructed images are inconsistent, they are adversarial examples that are effective to attack the targeted DNN model. From this perspective, the reason for this kind of adversarial attack is the inconsistency of latent features and input images. In other words, the mapping learned by the DNN model is not robust enough. There are still fundamental blind spots in the targeted DNN model~\cite{Goodfellow2014Exp}. 

\subsection{Adversarial Defence}
As we discussed above, one of the reasons for attack existence may be the existence of fundamental blind spots in the targeted DNN model. Is there any solution for the DNN model to avoid being attacked to some extent? 

In this paper, we adopt adversarial training to alleviate this problem. For adversarial training, the adversarial loss function can be rewritten as follows:
\begin{equation}
    \Tilde{J} = \alpha J\left(\theta; x, y\right) + (1-\alpha)J\left(\theta; \hat{G}\left(z+\epsilon \beta d\right),y\right)
    \label{eqa:loss}
\end{equation}
where $J(\theta; x,y)$ is the original loss function for the DNN classifier; $\alpha$ is the weighted factor to control importance of two parts, usually set as 0.5; $\hat{G}(z+\epsilon \beta d)$ is the generated adversarial example; $z=\phi(x)$, the latent features.

Through adversarial training, we fine-tune the targeted DNN model on the generated adversarial examples. We will later empirically show that, this fine-tuned DNN model is not only robust against attacks targeted on the classifier with the same structure as our feature extractor, but also effective to defend some other attacks targeted on other unknown and different classifiers. As shown in Fig.~\ref{fig:flowchat} (c), the new model is able to predict labels of adversarial examples correctly after adversarial training. The fine-tuned DNN model is more robust than the original one. 

\section{Experiments}
\label{sec:exp}

To evaluate the effectiveness of our proposed model, we conduct extensive experiments on MNIST~\cite{LecunMnist}, SVHN~\cite{netzer2011reading}, and CelebA~\cite{liu2015faceattributes}. We attempt to generate adversarial examples to attack the targeted DNN model, i.e., the DNN classifiers, show the mechanism of attack from the perspective of data space and defend different types of attacks with adversarially trained classifiers using our LAD. Specifically, in section \ref{exp:A}, we show the performance of our trained generator to make sure it captures the mapping between latent space and input space.  Generated adversarial examples by other attack methods and our LAD are compared in section \ref{exp:adv_sample}. Understanding attack mechanism may help defend various attacks in a uniform method. Therefore, we try to understand the attack progress in section \ref{exp:understand}. Section \ref{exp:defence} presents the defence results on MNIST, SVHN, and CelebA dataset. 

Note that, we use DNN models with different architectures and depths, e.g., LeNet~\cite{lecun1995learning} model, shallow and deep VGG model~\cite{simonyan2014very} as the targeted classifiers on the three datasets, respectively. Apart from these models, we also use  some unknown DNN classifiers whose model details are unknown to the service provider in model sharing scenarios, as the targeted classifiers. After adversarial training on generated adversarial examples, we also verify the effectiveness of the adversarially trained classifiers to defend against different types of attacks, including weak and powerful attacks, targeted and untargeted attacks, and attacks targeted on known and unknown classifiers. 

\subsection{Quality of Generated Examples}
\label{exp:A}
We first validate the performance of our trained generator in terms of its ability to capture the mapping between the latent space and the input space. Specifically, we trained a generator on the MNIST dataset using latent features, whose dimension is $500$, from the second-to-last layer of LeNet. In Appendix Table~\ref{app:mnist_g}, all the components of the generator architecture except for activation functions are listed. After the last convolutional layer, a sigmoid activation function is added and the loss function used is mean square error (MSE): 
$\ell(x, \hat{x}) = \frac{1}{n}\sum_{i=1}^n \left( x_i - \hat{x}_i \right)^2$.

\begin{figure}[tbp]
    \centering
    \includegraphics[width=1.0\columnwidth]{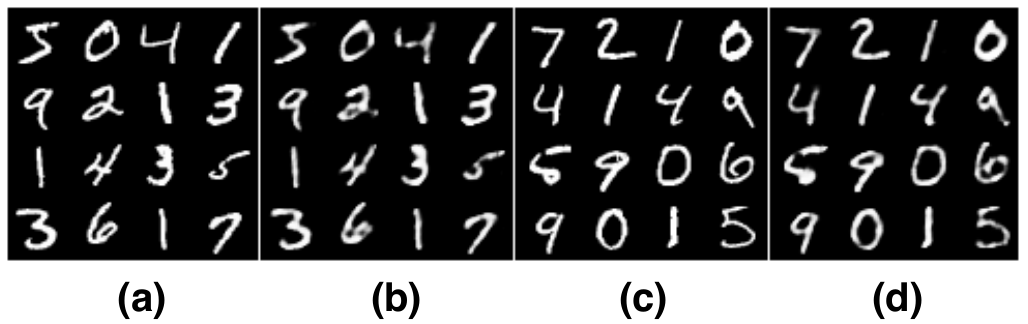}
    \caption{Reconstructed images of our generator trained on MNIST. (a) and (c) depict original training and test images, whereas (b) and (d) show generated training and test images.}
    \label{fig:gene_img}
\end{figure}

We evaluated our generator through quantitative and qualitative results. The training loss on training dataset after 1000 epoch and the test loss over test dataset are $0.00757$ and $0.00765$, respectively. Fig.~\ref{fig:gene_img} shows reconstructed images using our generator trained on MNIST. In detail, Fig.~\ref{fig:gene_img}(a) and ~\ref{fig:gene_img}(c) are original training and test images, while ~\ref{fig:gene_img}(b) and ~\ref{fig:gene_img}(d) are generated training and test images, which are very similar to the original images. From both quantitative and qualitative results, we can conclude that, the trained generator is able to capture the mapping between the latent space and the input space.

\subsection{Diversity of Adversarial Examples}
\label{exp:adv_sample}

We compare adversarial examples generated by our LAD model and other two attack methods (FGSM and JSMA) on MNIST. For our method, we used the trained generator to generate adversarial examples against the LeNet model. Latent features from the second-to-last fully connected layer in LeNet are used to train an SVM model which yields the boundary norm for generation. Each extracted latent feature is changed by adding perturbations with different strength ($\epsilon$). Finally, changed latent features are fed into the trained generator to generate adversarial examples. For FGSM and JSMA, we generated adversarial samples through \textit{cleverhans}~\cite{papernot2016technical}.

\begin{figure}[tbp]
    \centering
    \includegraphics[scale=1.1]{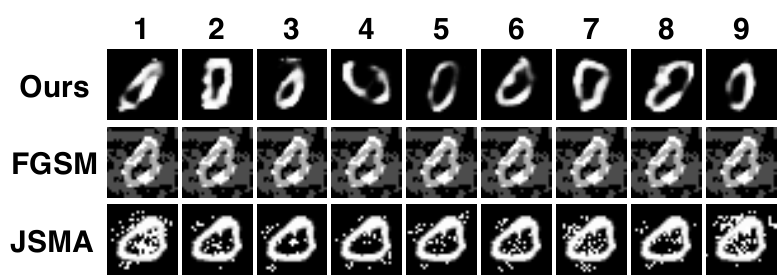}
    \caption{Adversarial examples generated by FGSM, JSMA, and our model, where the topmost number indicates class label. }
    \label{fig:our_AdSm}
\end{figure}

\paragraph{Variety of Adversarial Examples} Fig.~\ref{fig:our_AdSm} shows example images generated by FGSM, JSMA, and our method. These images are generated using different perturbations. Compared with adversarial examples generated by FGSM and JSMA, our LAD model is able to generate non-blurry images. It is worth noting that, our model generates a more diverse set of distinct examples, whereas the examples generated from FGSM and JSMA are always noisy images of repeating patterns. This is because our model generates these examples by modifying latent features rather than slightly altering the original images in the input space. 

\begin{figure}[htbp]
    \centering
    \includegraphics[scale=0.35]{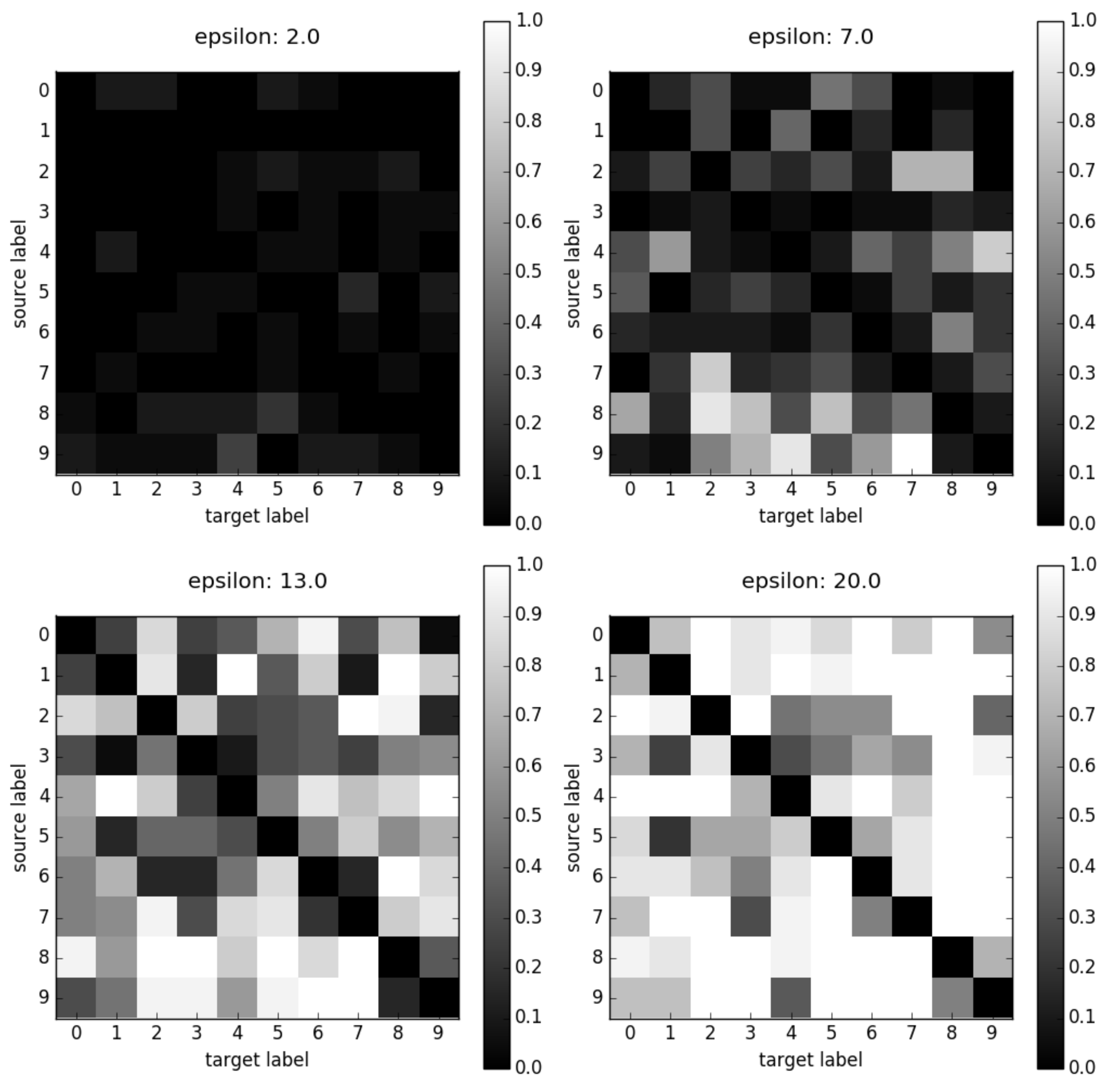}
    \caption{Attack success rates under different perturbations. The white color indicates a higher attack success rates.}
    \label{fig:our_AtRate}
\end{figure}

\paragraph{Attack Success Rate} Generated adversarial examples by our model are effective in attacking the pre-trained LeNet model. As shown in Fig.~\ref{fig:our_AtRate}, the attack success rate increases, as strength of perturbation $\epsilon$ increases. In particular, when $\epsilon$ is equal to $20.0$, attack success rates are very close to 1. This means that the generated adversarial examples by our LAD model are hard to predict correctly for the originally pre-trained classifier, so that these examples would be beneficial to improve the model.

\subsection{Understanding Attacks}
\label{exp:understand}

Why prediction models are easily attacked by generated adversarial examples? To answer this question, we analyze the robustness of prediction models against two types of attacks from the perspective of data space. Understanding the attack mechanism would be beneficial to find solutions to improve the adversarial robustness of prediction models. The adversarial examples in this part are generated through our LAD model.

\begin{figure}[tbp]
    \centering
    \includegraphics[scale=1.12]{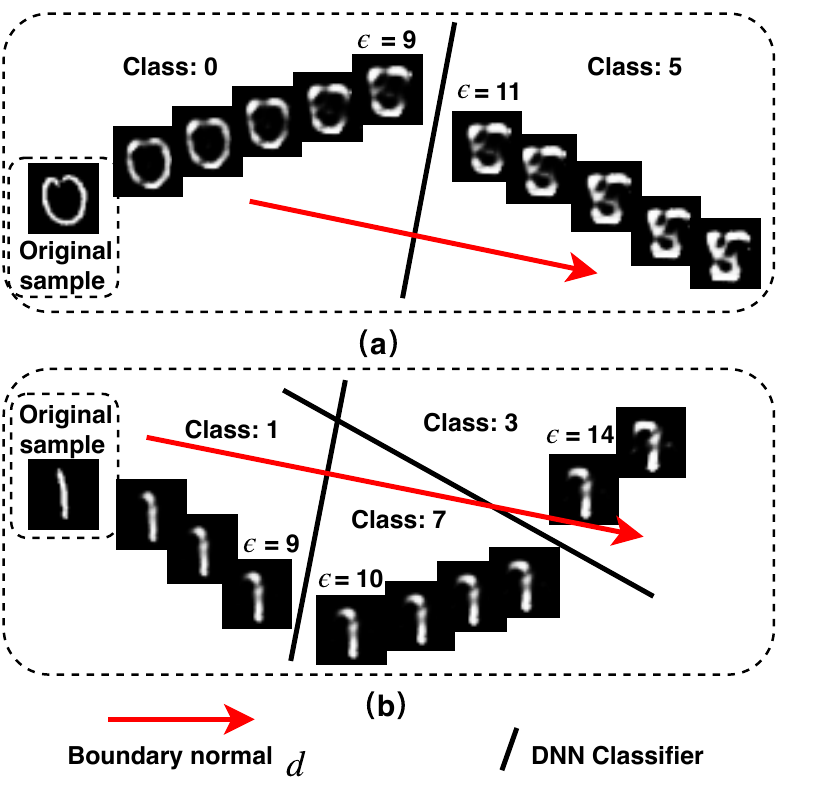}
    \caption{Transition of targeted attacks and polymorphism of attacks. The number on top of an image is the value of perturbation($\epsilon$).}
    \label{fig:boudary}
\end{figure}

\paragraph{Transition of Targeted Attacks} From classification perspectives, examples close to the decision boundary are more likely to be misclassified by a classifier. This also means, it is relatively easier to form a target attack for those examples near the decision boundary, because adding very small perturbations can change their predicted labels. As shown in Fig.~\ref{fig:boudary} (a), the original sample is apparently a digit $0$. When we increase the value of perturbation($\epsilon$) added to the corresponding latent features of the original sample along the boundary normal, the prediction of generated examples changes from $0$ to $5$. The two examples near the decision boundary, generated with $\epsilon = 9$ and $\epsilon = 11$, are inherently ambiguous, even making humans difficult to make a decision. If we add these ambiguous adversarial examples with labels to the training dataset, it would be helpful to improve the adversarial robustness of classifiers.


\paragraph{Polymorphism of Attacks} In another scenario where the data space is mixed with similar classes, if we generate examples in the mixed space, it is difficult for a classifier to give a correct label. For example, in Fig.~\ref{fig:boudary} (b), the label of the original sample is $1$. From the latent features of the same original sample, we generated examples with different perturbations along the boundary normal from class 1 to class 7. We found the labels of generated examples included not only class 1 and 7, but also class 3. Furthermore, DNN classifier predicted the two examples, generated with $\epsilon = 10$ and $\epsilon = 14$, as class 7 and class 3, while humans are more likely to classify them as class 1 and class 7. For these similarly adversarial examples, more and more examples with labels would help classifiers to be robust.

\subsection{Latent Adversarial Defence}
\label{exp:defence}
As discussed in the last section, generated adversarial examples, especially the one near decision boundary, would be beneficial to improve robustness of classifiers. Considering this, we fine-tuned the pre-trained classifier by augmenting the training dataset with the generated adversarial examples to improve model adversarial robustness against different attacks. Experiments were carried to prove effectiveness of our proposed model to defend against our own attacks as well as other types of attacks.

Experimental results on MNIST, SVHN, and CelebA are presented in this section. We conducted experiments to defend attacks targeted on known classifier, e.g., LeNet, shallow and deep VGG, as well as unknown classifiers for our LAD model. The attacks targeted on unknown classifiers refer to that the adversarial examples are generated through some attack methods to attack the classifiers which are unknown to our LAD model, i.e., the service provider in model sharing scenarios. If the adversarially trained classifiers through our LAD method are also able to defend the attacks targeted on unknown classifiers, it means our LAD method is effective to improve the adversarial robustness of classifiers even against unknown attacks. Note that, the adversarial examples from FGSM, JSMA, CW, and PGD were generated using adversarial examples library \textit{cleverhans}~\cite{papernot2016technical}, while the results from GA were generated by code in~\cite{song2018constructing}.

\begin{figure}[tbp]
    \center
    \includegraphics[scale=0.47]{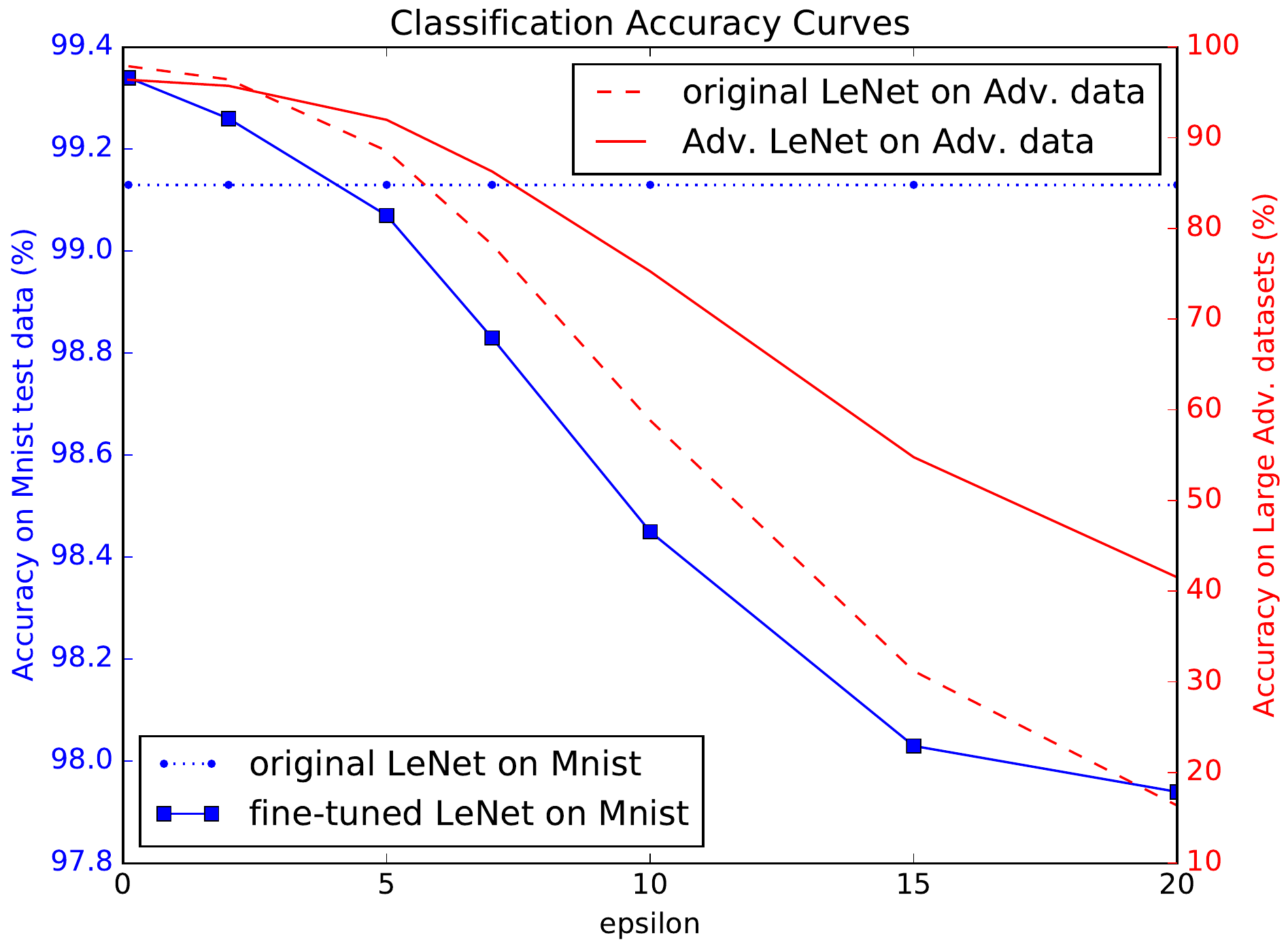}
    \caption{Classification accuracy of original LeNet and adversarially trained LeNet on MNIST test dataset and \textit{Large Adv. datasets} with different epsilon values.}
    \label{fig:acc_eps}
\end{figure}

\subsubsection{\textbf{Experiments on MNIST}}
In this part, we show the experimental results on MNIST. LeNet model is set as the shared classifier, i.e., the targeted classifier for attack. MNIST test dataset includes 10,000 legitimate examples, part of which are used in our defence experiments.

\paragraph{Robustness of Adversarially Trained Classifier} To adversarially train the pre-trained LeNet, we randomly selected 50 images per class from MNIST test dataset to generate 4500 adversarial examples for each strength of perturbations [0.1, 2.0, 5.0, 7.0, 10.0, 15.0, 20.0], which resulted in a dataset called \textit{Large Adv. datasets}. This dataset with different perturbations was separately used to fine-tune LeNet. We conducted classification task on clean MNIST test dataset using these fine-tuned LeNet. The results are given in Fig.~\ref{fig:acc_eps}, colored in blue(lines with square). It is clear to observe: 1). As $\epsilon$ increases, the classification accuracy of the fine-tuned model decreases. 2). When $\epsilon=0.1$, the fine-tuned LeNet outperforms original pre-trained LeNet. The reason for improved performance is that our proposed model enriches the data space with more unseen examples through boundary-guided generation. As such, the fine-tuned model is more robust. However, when $\epsilon$ continues to increase, the classification accuracy of adversarially fine-tuned LeNets decreases. This is because, the larger $\epsilon$ is, more ambiguous examples are generated. 

To further verify the effectiveness of defence using our LAD model, we adversarially trained the LeNet on mixed adversarial examples, which were generated with perturbations [0.1, 1.6, 3.2, 5.5, 10.0, 20.0] and are different from \textit{Large Adv. datasets}. We refer to this dataset as \textit{Small Adv. datasets} and the adversarially trained LeNet as \textit{Adv. LeNet}. From Fig.~\ref{fig:acc_eps}, the red part(lines without square), 
we can see that, our LAD fine-tuned model yields higher classification accuracy on adversarial examples than originally trained LeNet.


\begin{table}[tbp]
\centering
\caption{Defending attacks targetd on LeNet: classification accuracy of original LeNet and \textit{Adv. LeNet} on adversarial dataset targeting LeNet.}
\begin{tabular}{ccc}
\toprule
Datasets & original LeNet & \textit{Adv. LeNet} \\
\midrule
Adv. dataset (FGSM) & 1.77\% & 21.67\% \\
Adv. dataset (JSMA) & 8.71\% & 39.89\% \\
Adv. dataset (PGD) & 0.6\% & 9.44\% \\
Adv. dataset (CW) & 98.14\% & 98.48\% \\
\bottomrule
\end{tabular}
\label{tab:attack}
\end{table}

\paragraph{Defending Attacks Targeted on LeNet} We also verify the effectiveness of our adversarially trained LeNet model on our generated adversarial examples to defend attacks from other methods, such as FGSM, JSMA, PGD, and CW, when they have full access to LeNet. For FGSM and PGD attack methods, 450 (per class) randomly selected images in MNIST test dataset were used to generate $4500$ adversarial examples, which was under untargeted attack condition. But for JSMA and CW, 500 legitimate examples (50 per class) were randomly sampled to generate 4500 adversarial examples, which was under targeted condition. Note that, for FGSM and PGD, we set perturbation as $0.3$; for CW, we use $l_2$ norm distance. Table~\ref{tab:attack} compares defence performance of the original and \textit{Adv. LeNet} against FGSM, JSMA, PGD, and CW. We can see that, the adversarially trained LeNet model (\textit{Adv. LeNet}) on our generated adversarial examples are also effective to defend attacks from FGSM and JSMA, with improved classification accuracy on two adversarial datasets from $1.77\%$ to $21.67\%$ and from $8.71\%$ to $39.89\%$, respectively. For the more powerful attack methods, our model is also able to improve the classification accuracy, from 0.6\% to 9.44\% on PGD Adv. dataset and from 98.14\% to 98.48\% on CW Adv. dataset.

\paragraph{Defending Unknown Attacks} We also conducted experiments to investigate whether our adversarially trained LeNet model is useful for defending attacks targeted on unknown DNN classifier (called \textit{black CNN}). For our LAD model, there is no information about the unknown classifier used to generate adversarial examples by state-of-the-art methods. Apart from FGSM, JSMA, PGD, and CW, we also used GA to generate adversarial examples targeted on \textit{black} CNN to attack our fine-tuned LeNet model. Likewise, for FGSM and PGD, we also set perturbation as 0.3; for CW, we still chose $l_2$ norm distance. For the CW and JSMA \textit{unknown} adversarial datasets, 500 legitimate examples (50 per class) were randomly sampled to generate 4500 adversarial examples, which was under targeted attack condition. For other methods, 4500 legitimate examples (450 per class) were used for 4500 adversarial examples generation, which was under untargeted attack condition. 
Results are reported in Table~\ref{tab:attack2}, where the \textit{FT-0.1} LeNet is our adversarially trained model on adversarial examples with $\epsilon=0.1$. On \textit{unknown} adversarial dataset generated by FGSM, PGD and CW, \textit{FT-0.1} LeNet improves the accuracy to $49.60\%$, $35.93\%$ and $99.38\%$. This manifests that our adversarially trained LeNet model can robustly defend against FGSM, PGD and CW attack. On \textit{unknown} adversarial dataset (JSMA) and \textit{unknown} adversarial dataset (GA), the accuracy of \textit{FT-0.1} LeNet slightly drops as compared with the original LeNet. This is probably because, adding 4500 adversarial examples might also introduce noises, making it difficult to further improve the classification performance, given that the original LeNet already achieves the accuracy of 93.91\% and 99.82\%, respectively.

\begin{table}[tbp]
\centering
\caption{Defending unknown attacks: classification accuracy of original LeNet and \textit{FT-0.1} LeNet on \textit{unknown} adversarial dataset.}
\begin{tabular}{ccc} 
\toprule
Datasets & original LeNet & \textit{FT-0.1} LeNet\\
\midrule

\textit{unknown} Adv. data (FGSM) & 41.71\% & 49.60\% \\
\textit{unknown} Adv. data (JSMA) & 93.91\% & 92.44\% \\
\textit{unknown} Adv. data (PGD) & 29.93\% & 35.93\% \\
\textit{unknown} Adv. data (GA) & 99.82\% & 99.72\% \\
\textit{unknown} Adv. data (CW) & 99.12\% & 99.38\% \\
\bottomrule
\end{tabular}
\label{tab:attack2}
\end{table}

\paragraph{Defence in Latent Space vs. Input Space} Why do we use latent features rather than input images to generate adversarial examples and conduct latent adversarial defence? Experiments were carried to prove better performance by our latent adversarial defence. FGSM, JSMA, PGD, and CW are four baseline methods, which generate adversarial examples by adding perturbations in input space, while our method and GA do this in latent space. We generated adversarial examples and fine-tuned the LeNet model on the individually adversarial datasets and then evaluated the classification accuracy on clean MNIST test dataset. From Table~\ref{tab:latent_attack}, we can see that our fine-tuned LeNet model on adversarial examples with $\epsilon=0.1$ (\textit{FT-0.1}) obtains the best classification results. Even though we fine-tune LeNet on our adversarial examples with other perturbations $\epsilon$, classification accuracy on MNIST test dataset of our method is also the best (refer to Fig.~\ref{fig:acc_eps}). This means our LAD model does not hurt the classification performance with defending attacks.

\begin{table}[tbp]
\centering
\caption{Classification accuracy on MNIST test dataset using original and fine-tuned LeNet models on different adversarial examples generated by different attack methods.}
\begin{tabular}{cc}  
\toprule
Trained models & Classification accuracy\\ 
\midrule
Original LeNet & 99.13\%\\
Fine-tuned LeNet on FGSM & 93.57\% \\
Fine-tuned LeNet on JSMA & 95.71\% \\
Fine-tuned LeNet on PGD & 90.98\% \\
Fine-tuned LeNet on CW & 98.71\% \\
Fine-tuned LeNet on GA & 97.23\% \\
Fine-tuned LeNet on our model & \textbf{99.34\%} \\
\bottomrule
\end{tabular}
\label{tab:latent_attack}
\end{table}

\subsubsection{\textbf{Experiments on SVHN}}
In this part, we present the experimental results on SVHN~\cite{netzer2011reading} dataset to prove the effectiveness of the proposed method. A model based on shallow VGG is set as the shared classifier, which is called svhnNet (see Appendix Table \ref{app:svhnNet}). Features whose dimension are 4096, from the fifth layer in svhnNet, are extracted for training generator. The generator (see Appendix Table \ref{app:svhngenerator}) is trained on SVHN training dataset including 73257 examples. After that, we randomly select some images from each class in SVHN test dataset to generate adversarial examples. Note that, the original svhnNet hereinafter refers to svhnNet trained on clean SVHN training dataset and fine-tuned svhnNet is adversarially trained on adversarial examples generated by LAD with different $\epsilon$ or by other attack methods.

\paragraph{Defending Attacks Targeted on svhnNet}
We show experimental results of our model on defending attacks targeted on svhnNet in this part. The model, svhnNet, is set as the targeted classifier. For FGSM and PGD, we set perturbation as 0.3 and used 4500 randomly selected legitimate examples (450 per class) to generate 4500 adversarial examples under untargeted attack condition. For JSMA, 500 legitimate examples (50 per class) were randomly sampled to generate 4500 adversarial examples under targeted attack condition. Table \ref{tab:svhnAattack} presents the classification result on adversarial examples generated by different attack methods. On adversarial dataset generated by FGSM, PGD, and JSMA, \textit{FT-0.5} svhnNet improves the accuracy from $17.20\%$, $30.40\%$, and $35.78\%$ to $21.60\%$, $32.56\%$ and $37.40\%$, respectively. We see from the table, attacks from FGSM, PGD, and JSMA targeted on svhnNet are defended against by our fine-tuned svhnNet model.

\begin{table}[htbp]
\centering
\caption{Defending attacks targeted on svhnNet: classification accuracy of original and \textit{FT-0.5 svhnNet} on adversarial dataset targeting svhnNet. The \textit{FT-0.5} svhnNet is our fine-tuned model on adversarial examples with $\epsilon=0.5$.}
\begin{tabular}{ccc}
\toprule
Datasets & Original svhnNet & \textit{FT-0.5} svhnNet \\
\midrule
Adv. dataset (FGSM) & 17.20\% & 21.60\%\\
Adv. dataset (PGD) & 30.40\% & 32.56\%\\
Adv. dataset (JSMA) & 35.78\% & 37.40\%\\
\bottomrule
\end{tabular}
\label{tab:svhnAattack}
\end{table}

\paragraph{Defending Unknown Attacks}
Experimental results are present in this part to prove our fine-tuned svhnNet model is useful for defending unknown attacks. Because this is unknown attack, we set another DNN classifier, different from svhnNet, as the targeted model, which is not known by our LAD method. All experiments settings were the same as those in the last part, \textit{defending attacks targeted on svhnNet}. Table \ref{tab:svhnattack2} shows the classification accuracy under unknown attacks. On \textit{unknown} adversarial dataset generated by FGSM, PGD and JSMA, \textit{FT-2.0} svhnNet improves the accuracy from $25.00\%$, $17.16\%$, and $34.04\%$ to $27.80\%$, $17.36\%$ and $34.96\%$, respectively. This manifests that our fine-tuned svhnNet model can robustly defend against FGSM, PGD, and JSMA unknown attacks.

\begin{table}[tbp]
\centering
\caption{Defending unknown attacks: classification accuracy of original and \textit{FT-2.0} svhnNet on \textit{unknown} adversarial dataset. The \textit{FT-2.0} svhnNet is our fine-tuned model on adversarial examples with $\epsilon=2.0$.}
\begin{tabular}{ccc} 
\toprule
    Datasets & Original svhnNet & \textit{FT-2.0} svhnNet\\
\midrule
\textit{unknown} Adv. dataset (FGSM) & 25.00\% & 27.80\%\\
\textit{unknown} Adv. dataset (PGD) & 17.16\% & 17.36\%\\
\textit{unknown} Adv. dataset (JSMA) & 34.04\% & 34.96\%\\
\bottomrule
\end{tabular}
\label{tab:svhnattack2}
\end{table}

\paragraph{Robustness of Adversarially Trained Classifier}
In this part, classification results on clean SVHN test dataset by different fine-tuned svhnNet are shown. We fine-tuned the svhnNet on adversarial datasets generated by FGSM, PGD, and JSMA that was targeted on svhnNet. In addition, svhnNet was also adversarially trained on our generated adversarial dataset. All the fine-tuned svhnNet were used to classify the clean SVHN test dataset. Table \ref{tab:svhnlatent_attack} presents the classification accuracy in details. We could find fine-tuned svhnNet on our adversarial dataset under $\epsilon = 0.05$ gets the best performance. Even the bad result of our model is still better than other methods'. This indicates the fine-tuned model by our method is more robust.

\begin{table}[htbp]
\centering
\caption{Classification accuracy on SVHN test dataset using original and fine-tuned svhnNet models using different adversarial examples. The \textit{FT-number} svhnNet is our fine-tuned model on adversarial examples with different $\epsilon$.}
\begin{tabular}{cc}  
\toprule
Trained models & Classification accuracy\\ 
\midrule

Original svhnNet & 93.85\%\\
Fine-tuned svhnNet on JSMA & 93.27\% \\
Fine-tuned svhnNet on PGD & 91.49\% \\
Fine-tuned svhnNet on FGSM & 91.32\% \\
FT-0.05 svhnNet on our model & \textbf{94.83}\% \\
FT-0.15 svhnNet on our model & 94.80\% \\
FT-0.5 svhnNet on our model & 94.58\% \\
\bottomrule
\end{tabular}
\label{tab:svhnlatent_attack}
\end{table}

\subsubsection{\textbf{Experiments on CelebA}}
We also conducted experiments on CelebA dataset and the results are claimed in this section. Our task is the classification of smile or non-smile for an input image. The training dataset is part of CelebA, including 19999 images. The test dataset consists of 2000 images randomly chose from the whole dataset. Because the size of original images in CelebA is 178$\times$218, we firstly pre-process the images to 128$\times$128 using DLIB~\cite{dlib}. We detect faces in images and crop them into square size. The classifier for this classification task is based on VGG net~\cite{simonyan2014very}. We fixed the convolution and pooling layers of VGG11 as the feature extractor and add more fully connected layers to reduce dimensions and get labels. We called this model as CelebANet (See Appendix Table \ref{app:CelebANet} for details) and the model trained on the clean CelebA training dataset is called original CelebANet. In addition, the architecture of boundary-guided generator for CelebA is shown in Appendix Table \ref{app:celebAgenerator}.

\paragraph{Defending Attacks Targeted on CelebANet}
On CelebA dataset, we also conducted experiment, defending attacks targeted on CelebANet from FGSM and PGD, to prove our model's effectiveness. For both attack methods, 2000 benign images in test dataset are used to generate 2000 adversarial examples, with the perturbation $\epsilon=0.3$. These adversarial examples constitute Adv. dataset (FGSM) and Adv. dataset (PGD). Analogously, we used our model to generate adversarial examples with perturbations $[0.1, 1.0, 2.0, 3.0]$, to form Adv. dataset (our model).
In this experiment, attack methods have full access to CelebANet. We classified the adversarial dataset by original CelebANet and fine-tuned CelebANet. As shown in Table \ref{tab:celebAWhiteattack}, the \textit{FT-0.1} CelebANet improves classification accuracy on Adv. dataset (our model $\epsilon=2.0$) from 81.05\% to 84.30\%. \textit{FT-0.1} CelebANet is the fine-tuned model on our adversarial examples with $\epsilon=0.1$). As for the slightly dropped accuracy on Adv. dataset (FGSM and PGD), the two factors that generating better adversarial examples on CelebA is harder and the attack methods know all the details of CelebANet model, make the defence by adversarial training more difficult.

\begin{table}[tbp]
\centering
\caption{Defending attacks targeted on CelebANet: classification accuracy on {\upshape Adv. dataset} by original CelebANet and \textit{FT-0.1} CelebANet on {\upshape Adv. dataset (our model $\epsilon=2.0$)}.}
\resizebox{0.5\textwidth}{8.8mm}{
\begin{tabular}{ccc}
\toprule
Datasets & original CelebANet & \textit{FT-0.1} CelebANet \\
\midrule
Adv. dataset (FGSM) & 52.75\% & 50.90\% \\
Adv. dataset (PGD) & 12.25\% & 11.75\% \\
Adv. dataset (our model $\epsilon=2.0$) & 81.05\% & 84.20\%\\
\bottomrule
\end{tabular}
}
\label{tab:celebAWhiteattack}
\end{table}

\paragraph{Defending Unknown Attacks}
Experiments are carried out to defend attacks targeted on other DNN models (called \textit{black CelebACNN}) and the results are present in this part. This \textit{black CelebACNN} can only access the training dataset of pre-processed CelebA images. We used FGSM and PGD as attack methods to attack \textit{black CelebACNN}. Likewise, we randomly selected 2000 legitimate examples to form CelebA test dataset to generate 2000 adversarial examples by the two attack methods, which formed \textit{unknown} Adv. dataset (FGSM) and \textit{unknown} Adv. dataset (PGD). The generated adversarial dataset was classified by the original CelebANet and the fine-tuned CelebANet. Results are reported in Table~\ref{tab:celebAattack2}, where the \textit{FT-0.1} CelebANet is our fine-tuned model on adversarial examples with $\epsilon=0.1$. We can see that the \textit{FT-0.1} CelebANet is able to defend this kind of attacks, with improving the classification accuracy on \textit{unknown} adversarial dataset of FGSM and PGD from $52.65\%$ to $52.8\%$, and from $13.90\%$ to $32.55\%$, respectively. This manifests that our fine-tuned model can robustly defend against these attacks. 

\begin{table}[tbp]
\centering
\caption{Defending unknown attacks: classification accuracy on \textit{black} {\upshape Adv. dataset} by original CelebANet and \textit{FT-0.1} CelebANet.}
\begin{tabular}{ccc} 
\toprule
Datasets & original CelebANet & \textit{FT-0.1} CelebANet\\
\midrule
\textit{unknown} Adv. data (FGSM) & 52.65\% & 52.8\% \\
\textit{unknown} Adv. data (PGD) & 13.90\% & 32.55\% \\
\bottomrule
\end{tabular}
\label{tab:celebAattack2}
\end{table}

\paragraph{Robustness of Adversarially Trained Classifier}
In order to verify our model's ability to improve  adversarial robustness of classifier, we compared the classification performance of different fine-tuned CelebANet on clean CelebA test datasets. We fine-tuned CelebANet on different adversarial datasets generated by different attack methods and our model. After that, we used these fine-tuned classifiers to classify the CelebA test dataset. From Table \ref{tab:celebAlatent_attack}, it is obvious that our model acquires the best result, and especially improves $8.45\%$ compared with the original CelebANet which is trained on clean CelebA training dataset. This proves that our fine-tuned model is more robust than others.

\begin{table}[htbp]
\centering
\caption{Classification accuracy on CelebA (smile and non-smile) test dataset using different fine-tuned CelebANet models on different adversarial datasets.}
\begin{tabular}{cc}  
\toprule
Trained models & Classification accuracy\\ 
\midrule
Original CelebANet & 91.40\%\\
Fine-tuned CelebANet on FGSM & 91.20\% \\
Fine-tuned CelebANet on PGD & 92.35\% \\
Fine-tuned CelebANet on our model($\epsilon=0.1$) & \textbf{99.85}\% \\
\bottomrule
\end{tabular}
\label{tab:celebAlatent_attack}
\end{table}

Overall, our fine-tuned model is robust against different types of attacks, even the attacks targeted on other unknown classifiers. The proposed LAD method is able to improve adversarial robustness of the classifier without sacrifice the classification performance. Therefore, our LAD model is very useful for training classifiers to improve its robustness against various kinds of attacks in sharing model scenarios, before the shared model is released.

\section{Conclusion} 
In this paper, we propose a novel method \textit{Latent Adversarial Defence} (LAD), which is based on decision boundary in latent space for generating adversarial examples and defending adversarial attacks. Our model can be used to generate high-quality and diverse adversarial examples and those examples are also effective to attack a DNN model. After adversarial training on our generated adversarial examples, the fine-tuned targeted DNN model is effective against different types of attacks, even the attacks targeted on other unknown classifiers. In model sharing scenarios, our LAD is able to improve adversarial robustness of shared classifiers before they are deployed as a public service. Later, we will extend our proposed model to generate adversarial examples based on multi-class decision boundary and on larger size image dataset, especially the generator to generate better quality examples on these larger natural images dataset, such as ImageNet. We will also try to give the theoretical guarantee to improve the adversarial robustness of classifiers.


%

\appendix
\begin{table}[htbp]
\caption{Architecture of boundary-guided generator for MNIST. \textit{Linear} indicates linear transformation; \textit{Conv\_Transpose} denotes transposed convolution; and \textit{Conv} represents convolution.}
\label{tab:generator}
\centering
\small
\begin{tabular}{ccc}
\toprule
Layers & Layer parameters\\
\midrule
Linear & input: 500, output: 50 $\times$ 4 $\times$ 4 \\
Conv\_Transpose &  kernel: 2 $\times$ 2, stride: 4 $\times$ 4\\
Conv &  kernel: 3 $\times$ 3, stride: 1 $\times$ 1\\
Conv\_Transpose &  kernel: 2 $\times$ 2, stride: 3 $\times$ 3\\
Conv &  kernel: 4 $\times$ 4, stride: 1 $\times$ 1\\
Conv &  kernel: 5 $\times$ 5, stride: 1 $\times$ 1\\
\bottomrule
\end{tabular}
\label{app:mnist_g}
\end{table}

\begin{table}[!ht]
\centering
\caption{The architecture of svhnANet. \textit{Conv} means convolution layer; \textit{Linear} indicates linear transformation; \textit{kernels} means number of kernels; \textit{kernel} means the dimension of kernel; \textit{stride} means the steps of convolutions.}
\begin{tabular}{cc}  
\toprule
Layers &  Layer Parameters\\ 
\midrule
Conv \& Maxpool & 64 kernels, kernel:3 x 3, stride: 1 x 1\\
\midrule
Conv \& Maxpool & 128 kernels, kernel:3 x 3, stride: 1 x 1\\
\midrule
Conv \& Maxpool & 256 kernels, kernel:3 x 3, stride: 1 x 1\\
\midrule
Conv \& Maxpool & 512 kernels, kernel:3 x 3, stride: 1 x 1\\
\midrule
Conv & 512 kernels, kernel:3 x 3, stride: 1 x 1\\
Linear \& ReLU & input: 512 $\times$ 4 $\times$ 4, output: 4096  \\
\midrule
Dropout & dropout rate: 0.5\\
Linear \& ReLU & input: 4096, output: 2048  \\
Linear \& ReLU & input: 2048, output: 512  \\
Linear & input: 512, output: 10\\
\bottomrule
\end{tabular}
\label{app:svhnNet}
\end{table}


\begin{table}[!h]
\caption{Architecture of boundary-guided generator for SVHN. \textit{Linear} indicates linear transformation; \textit{Conv\_Transpose} denotes transposed convolution; and \textit{Conv} represents convolution; \textit{BN} represents batch normalization;  \textit{kernels} means number of kernels; \textit{kernel} means the dimension of kernel; \textit{stride} means the steps of convolutions.}
\centering
\small
\begin{tabular}{ccc}
\toprule
Layers & Layer parameters\\
\midrule
Linear & input: 4096, output: 512 $\times$ 2 $\times$ 2 \\
Conv \& BN \& ReLU &  kernels: 512, kernel: 3 $\times$ 3, stride: 1\\
Conv\_Transpose \& BN & kernels: 512, kernel: 2 $\times$ 2, stride: 1\\
\midrule
Conv \& BN \& ReLU &  kernels: 512, kernel: 3 $\times$ 3, stride: 1\\
Conv\_Transpose \& BN & kernels: 512, kernel: 2 $\times$ 2, stride: 1\\
\midrule
Conv \& BN \& ReLU &  kernels: 256, kernel: 3 $\times$ 3, stride: 1\\
Conv\_Transpose \& BN & kernels: 256, kernel: 2 $\times$ 2, stride: 1\\
\midrule
Conv \& BN \& ReLU &  kernels: 128, kernel: 3 $\times$ 3, stride: 1\\
Conv\_Transpose \& BN & kernels: 128, kernel: 2 $\times$ 2, stride: 1\\
\midrule
Conv \& BN \& ReLU &  kernels: 64, kernel: 3 $\times$ 3, stride: 1\\
Conv \& Tanh &  kernels: 3, kernel: 1 $\times$ 1, stride: 1\\
\bottomrule
\end{tabular}
\label{app:svhngenerator}
\end{table}

\begin{table}[!h]
\centering
\caption{The architecture of CelebANet for classifying CelebA dataset. \textit{Linear} indicates linear transformation; Feature Extractor is the part of classifier used for extracting features to construct boundary.}
\begin{tabular}{cc}  
\toprule
Layers &  Layer Parameters\\ 
\midrule
Feature Extractor & All Conv and Pooling Layers of VGG11\\
Linear & input: 512 $\times$ 4 $\times$ 4, output: 4096  \\
ReLU \& Dropout & dropout rate: 0.5\\
Linear & input: 4096, output: 2048  \\
ReLU \& Dropout & dropout rate: 0.5\\
Linear & input: 2048, output: 1\\
\bottomrule
\end{tabular}
\label{app:CelebANet}
\end{table}

\begin{table}[!h]
\caption{Architecture of boundary-guided generator for CelebA. \textit{Linear} indicates linear transformation; \textit{Conv\_Transpose} denotes transposed convolution; and \textit{Conv} represents convolution; \textit{BN} represents batch normalization.  \textit{kernels} means number of kernels. \textit{kernel} means the dimension of kernel. \textit{stride} means the steps of convolutions.}
\centering
\small
\resizebox{0.49\textwidth}{3cm}{
\begin{tabular}{cccc}
\toprule
Layers & Layer parameters & Repeat\\
\midrule
Linear & input: 4096, output: 512 $\times$ 4 $\times$ 4 & 1\\
Conv\_Transpose \& BN & kernels: 512, kernel: 2, stride: 1 & 1\\
Conv \& BN \& ReLU &  kernels: 512, kernel: 3, stride: 1 & 3\\
\midrule
Conv\_Transpose \& BN & kernels: 512, kernel: 2, stride: 1 & 1\\
Conv \& BN \& ReLU &  kernels: 512, kernel: 3, stride: 1 & 3\\
\midrule
Conv\_Transpose \& BN & kernels: 256, kernel: 2, stride: 1 & 1\\
Conv \& BN \& ReLU &  kernels: 256, kernel: 3, stride: 1 & 3\\
\midrule
Conv\_Transpose \& BN & kernels: 128, kernel: 2, stride: 1 & 1\\
Conv \& BN \& ReLU &  kernels: 128, kernel: 3, stride: 1 & 2\\
\midrule
Conv\_Transpose \& BN & kernels: 64, kernel: 2, stride: 1 & 1\\
Conv \& BN \& ReLU &  kernels: 64, kernel: 3, stride: 1 & 2\\
\midrule
Conv \& Tanh &  kernels: 3, kernel: 1, stride: 1 & 1\\
\bottomrule
\end{tabular}
}
\label{app:celebAgenerator}
\end{table}




\ifCLASSOPTIONcaptionsoff
  \newpage
\fi



\bibliographystyle{IEEEtran}
\bibliography{TNNLS}
\end{document}